\def\year{2022}\relax
\documentclass[letterpaper]{article} 
\usepackage{aaai22}  
\usepackage{times}  
\usepackage{helvet}  
\usepackage{courier}  
\usepackage[hyphens]{url}  
\usepackage{graphicx} 
\usepackage{natbib}  
\usepackage{caption} 
\DeclareCaptionStyle{ruled}{labelfont=normalfont,labelsep=colon,strut=off} 
\usepackage{algorithm}
\usepackage{algorithmic}

\usepackage{caption}
\usepackage{wrapfig}
\usepackage{color}
\usepackage{xcolor}
\usepackage{algorithm}
\usepackage{algorithmic}

\usepackage{makecell}
\usepackage{booktabs,multirow,rotating}

\usepackage{colortbl}

\usepackage{ragged2e}

\usepackage{float}
\usepackage{subcaption}
\usepackage{amsmath}
\usepackage{amssymb}
\usepackage{diagbox}

\usepackage{bbding}

\usepackage{newfloat}
\usepackage{listings}
\floatstyle{ruled}
\newfloat{listing}{tb}{lst}{}
\floatname{listing}{Listing}
\title{Pixel is All You Need: Adversarial Trajectory-Ensemble  Active Learning for Salient Object Detection}
\author{
    Zhenyu Wu\textsuperscript{\rm 1},  Lin Wang\textsuperscript{\rm 2}, Wei Wang\textsuperscript{\rm 3}, Qing Xia\textsuperscript{\rm 4},
     Chenglizhao Chen\textsuperscript{\rm 5}\footnote{Corresponding Author: Chenglizhao Chen, cclz123@163.com}, 
     Aimin Hao\textsuperscript{\rm 1,6}, Shuo Li\textsuperscript{\rm 7}
}
\affiliations{
    \textsuperscript{\rm 1}State Key Laboratory of Virtual Reality Technology and Systems, Beihang University \\ 
    \textsuperscript{\rm 2}School of Transportation Science and Engineering, Beihang University \\ 
    \textsuperscript{\rm 3}Harbin Institute of Technology (Shenzhen), \textsuperscript{\rm 4}SenseTime Research\\
    \textsuperscript{\rm 5}China University of Petroleum (East China), \textsuperscript{\rm 6}Peng Cheng Laboratory, \textsuperscript{\rm 7}Case Western Reserve University
}

\usepackage{bibentry}

\begin{document}

\maketitle

\begin{abstract}
Although weakly-supervised techniques can reduce the labeling effort, it is unclear whether a saliency model trained with weakly-supervised data (e.g., point annotation) can achieve the equivalent performance of its fully-supervised version. This paper attempts to answer this unexplored question by proving a hypothesis: there is a point-labeled dataset where saliency models trained on it can achieve equivalent performance when trained on the densely annotated dataset. To prove this conjecture, we proposed a novel yet effective adversarial trajectory-ensemble active learning (ATAL). Our contributions are three-fold:  1) Our proposed adversarial attack triggering uncertainty can conquer the overconfidence of existing active learning methods and accurately locate these uncertain pixels. {2)} Our proposed trajectory-ensemble uncertainty estimation method maintains the advantages of the ensemble networks while significantly reducing the computational cost. {3)} Our proposed relationship-aware diversity sampling algorithm can conquer oversampling while boosting performance.
Experimental results show that our ATAL can find such a point-labeled dataset, where a saliency model trained on it obtained $97\%$ -- $99\%$ performance of its fully-supervised version with only ten annotated points per image. 
\end{abstract}

\section{Introduction}
Salient object detection (SOD) aims to segment the most attractive regions in an image according to the human perception system. 
Recently, deep learning based SOD methods \cite{Ji_2021_CVPR,Fan_2021_CVPR,Sun_2021_CVPR,Liu_2021_ICCV,Tang_2021_ICCV,Gu_2021_ICCV,Zhou_2021_ICCV} have achieved great success with the help of well-annotated large-scale datasets. Unfortunately, it is a prohibitive cost to annotate a large amount of pixel-wise data. 

Weakly-supervised salient object detection (WSOD) methods have been proposed to alleviate the dependency on pixel-wise data. Existing WSOD methods ~\cite{zeng2019multi,li2018weakly,wang2017learning} utilize weak annotation, such as image-level label and scribble annotation, which is easier to collect than a densely annotated label. More recently, \cite{PSOD_aaai2022} proposed a point-supervised SOD framework with the help of edge information. Despite the substantial progress, it is unclear whether a saliency model trained with weakly-supervised data (e.g., point-annotation) can achieve the equivalent performance of its fully-supervised version (i.e., pixel-wise annotation).

\begin{figure}[t]
\centering
\includegraphics[width=0.41\textwidth]{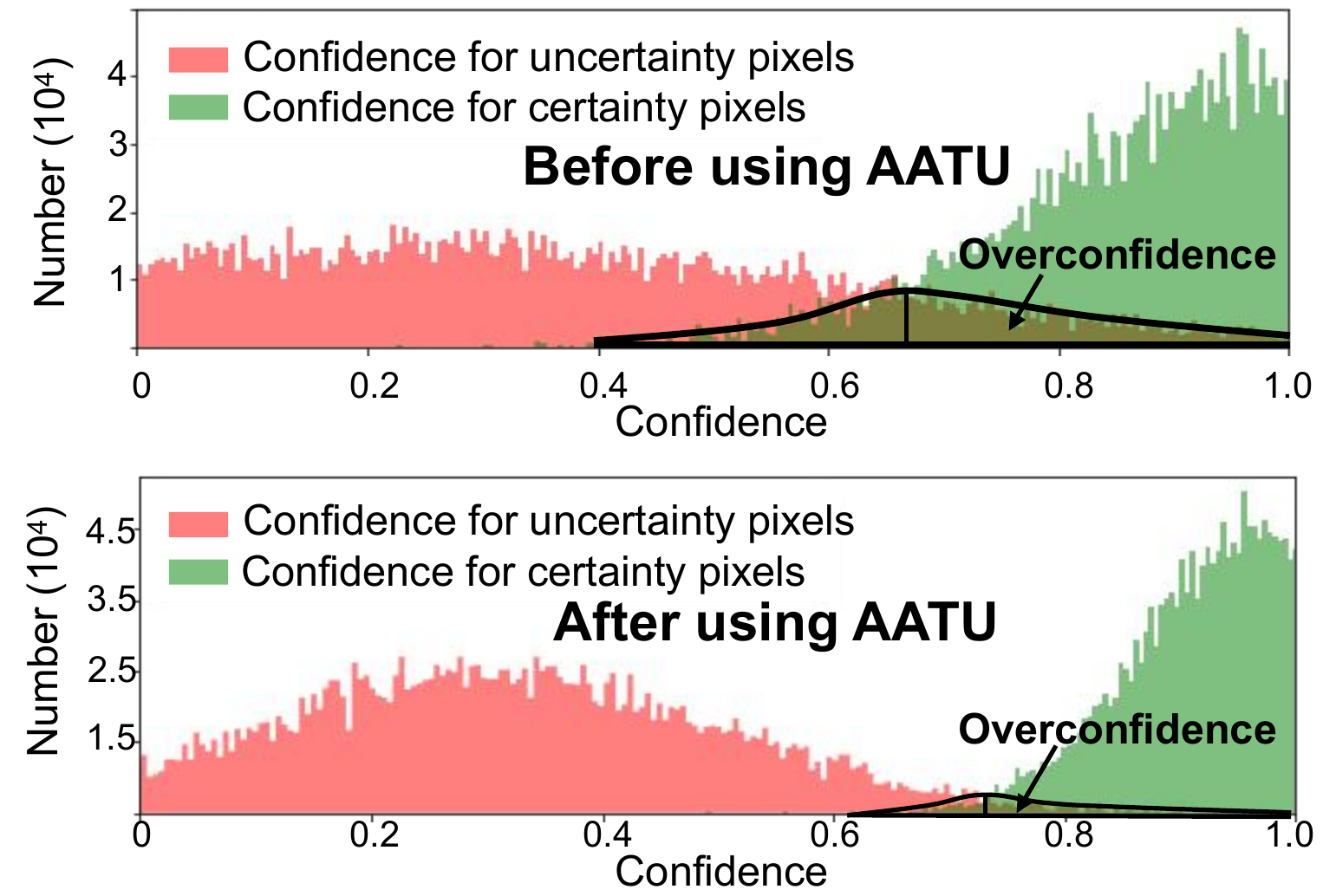} \vspace{-0.7em}
\caption{Our AATU can conquer the overconfidence issue of existing active learning methods and accurately identify these uncertain pixels.}
\label{fig:overconfidence}
\end{figure}

\begin{figure}[t]

\centering
\resizebox{0.45\textwidth}{!}{
\includegraphics[width=1\textwidth]{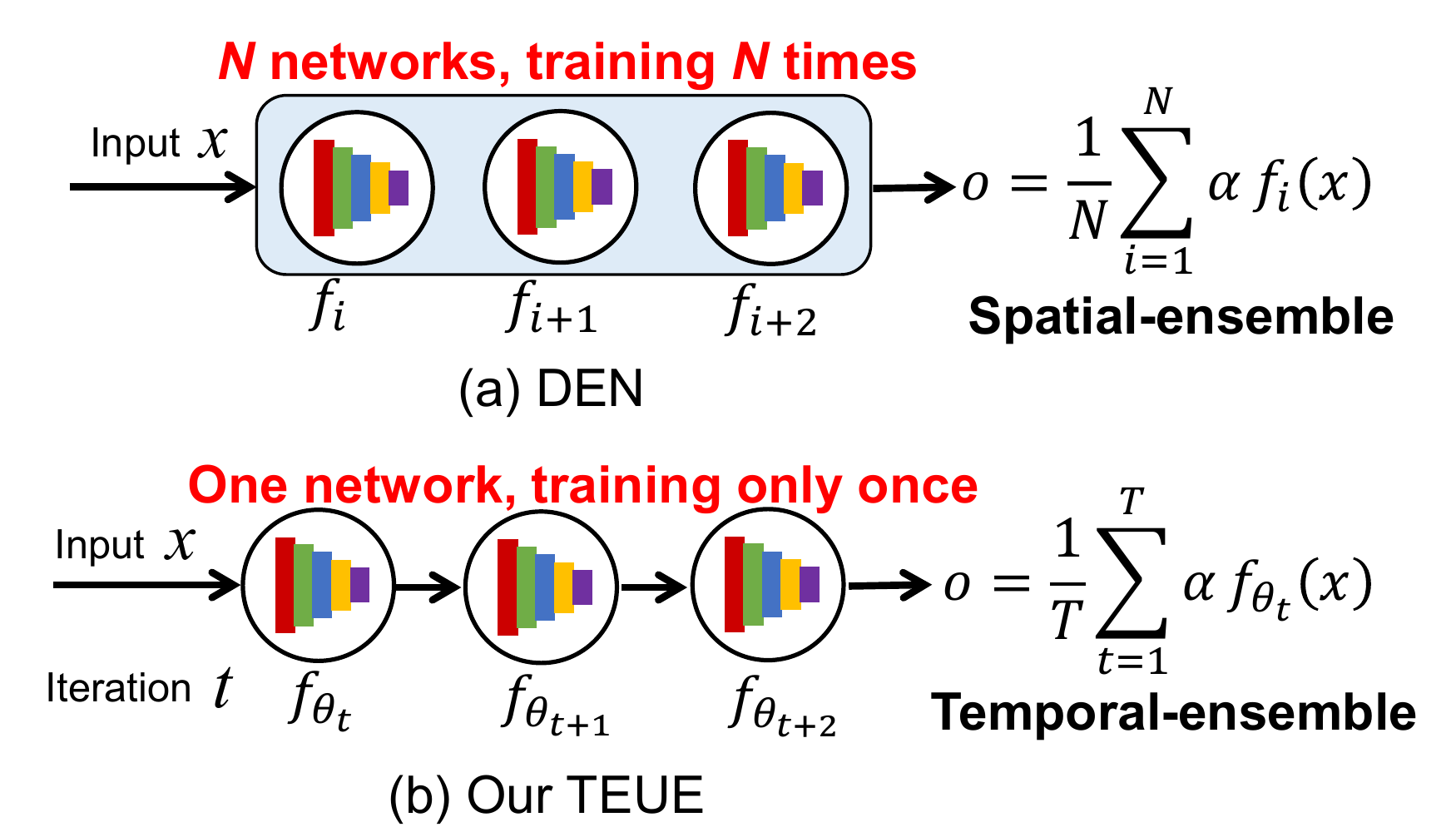}
}\vspace{-0.5em}
\caption{Unlike traditional DEN, which requires training $N$ networks, our TEUE requires one network and training only once, compressing the computational overhead to $1/N$.}
\label{fig:differences}
\end{figure}

In this paper, we answer this unexplored question by proving a hypothesis: 
\textit{there is a point labeled dataset where saliency models trained on it could achieve equivalent performance when trained on the densely annotated dataset.}
This conjecture, if it is true, has rather promising practical significance--it suggests that the pixel-wise annotation is in fact unnecessary, freeing humans from the heavy burden of labeling pixel-wise data.
While finding such a point-labeled dataset is non-trivial, there are two main challenges: 
\textbf{1)} In standard active learning (AL), uncertain pixels are defined as the low confidence pixels to the current model that has been trained over the present labeled set. However, our experiments revealed that existing AL methods are prone to produce ``overconfident'' predictions for uncertain pixels (see the 1st row of Fig. \ref{fig:overconfidence}), resulting in inaccurate estimation. 
\textbf{2)} Despite deep ensemble networks (DEN) \cite{lakshminarayanan2017simple,franchi2020tradi,zaidi2021neural} is the most effective methodology for uncertainty estimation, it introduces prohibitive training cost, which is mainly incurred by training $N$ networks repeatedly (see Fig. \ref{fig:differences}a).
For instance, compared to the vanilla training baseline, the NES \cite{zaidi2021neural} increases the total computational cost by 10$\times$.

To address these challenges, we proposed a novel yet effective Adversarial Trajectory-ensemble Active Learning (ATAL), which can accurately identify these uncertain pixels and run cheaply as the vanilla training baseline. 
\textbf{First}, instead of focusing on developing a new regularizer to mitigate the overconfidence of existing softmax AL methods, we argue that continued research progress in uncertainty estimation requires new insights. In this paper, we take drastically different views, and propose a surprisingly effective technique, Adversarial Attack Triggering Uncertainty (AATU), to explicitly identify these uncertain pixels by injecting the adversarial perturbation into the input image. In contrast to previous methods, AATU allows us to comprehensively evaluate the uncertainty of each pixel and conquer the overconfidence of modern AL methods (see the 2nd row of Fig. \ref{fig:overconfidence}). 
\textbf{Second}, by rethinking the standard deep ensemble networks and their variants, we proposed a temporal-ensemble model called Trajectory-Ensemble Uncertainty Estimation (TEUE) by aggregating network weights of the history model on the optimization path during the process of training. Unlike the previous spatial-ensemble techniques (see Fig. \ref{fig:differences}a), which require $N$ networks and repeat training $N$ times, our TEUE only requires one network and training only once (see Fig. \ref{fig:differences}b). Besides, our proposed Relationship-aware Diversity Sampling (RDS) can alleviate the oversampling issues and further boost the performance by considering the relationship among these sampling pixels.

Our ATAL has several desirable properties: 
\textbf{1) High Performance.} Our ATAL can achieve averagely $97\%$ -- $99\%$ performance of its fully-supervised version with only ten annotated pixels per image (see Table. \ref{tab:main_results}). In addition, our proposed method also outperforms existing WSOD models by a large margin.
\textbf{2) Low Computational Cost.} Compared to traditional $N$-ensemble networks, our ATAL can compress the computational overhead to $1/N$.
\textbf{3) Generalization.} Our ATAL can be easily integrated into existing SOD models as a plug-and-play module. Besides, a point labeled dataset selected by one SOD model can be used to train other SOD models well.
In summary, our main contributions are:

\begin{itemize}
\item  For the first time, we demonstrated that a saliency model trained with a point-labeled dataset could achieve equivalent performance trained on the pixel-wise dataset.
\item Our AATU provides a new insight for uncertainty estimation and identifies these uncertain pixels by injecting the adversarial attacks into the input image, which can alleviate the overconfidence of modern AL methods and accurately locate these uncertain pixels.
\item Our TEUE reduces the computational overheads stemming from repeatedly network training while maintaining the advantages brought by the ensemble networks.
\item Our proposed RDS algorithm can alleviate the oversampling issues and further boost the performance by considering the relationship among these sampling pixels.
\end{itemize}

\section{Related Works}

\noindent{\textbf{Weakly-supervised SOD.} With recent advances in weakly-supervised learning, a few existing works exploit the potential of training saliency model on image-level~\cite{zeng2019multi,li2018weakly,wang2017learning,piao2021mfnet} and scribble-level~\cite{yu2021structure,zhang2020weakly,zhang2020learning} to relax the dependency of manually annotated pixel-level masks. These approaches follow the same technical route, i.e., producing the initial saliency maps with image-level labels and then refining them via iterative training.  
Recently, point annotation was proposed in~\cite{PSOD_aaai2022}, but it requires extra data information (e.g., edge) to recover integral object structure.
\textbf{Differences.} Distinct from all these works, our work attempt to demonstrate that a saliency model trained on a point-labeled dataset can achieve equivalent performance when trained on the pixel-wise dataset.


\noindent{\textbf{Deep Active Learning.} Active learning is a set selection problem that aims to determine the most informative subset given a labeling budget. It has been successfully used as a method for reducing labeling costs. Recently, deep active learning has been attracting increasingly more attention in many areas, including classification \cite{beluch2018power,gal2017deep,krishnamurthy2017active,gao2020consistency}, object detection \cite{yoo2019learning,yuan2021multiple,aghdam2019active,Choi_2021_ICCV},  semantic segmentation \cite{kim2021task,siddiqui2020viewal,dai2020suggestive,yang2017suggestive}.  \textbf{Differences.} In contrast to previous methods, our ATAL can conquer the overconfidence of modern AL methods and accurately locate these uncertain pixels, significantly reducing the computational cost.

\begin{figure*}[t]
\centering
\includegraphics[width=0.9\textwidth]{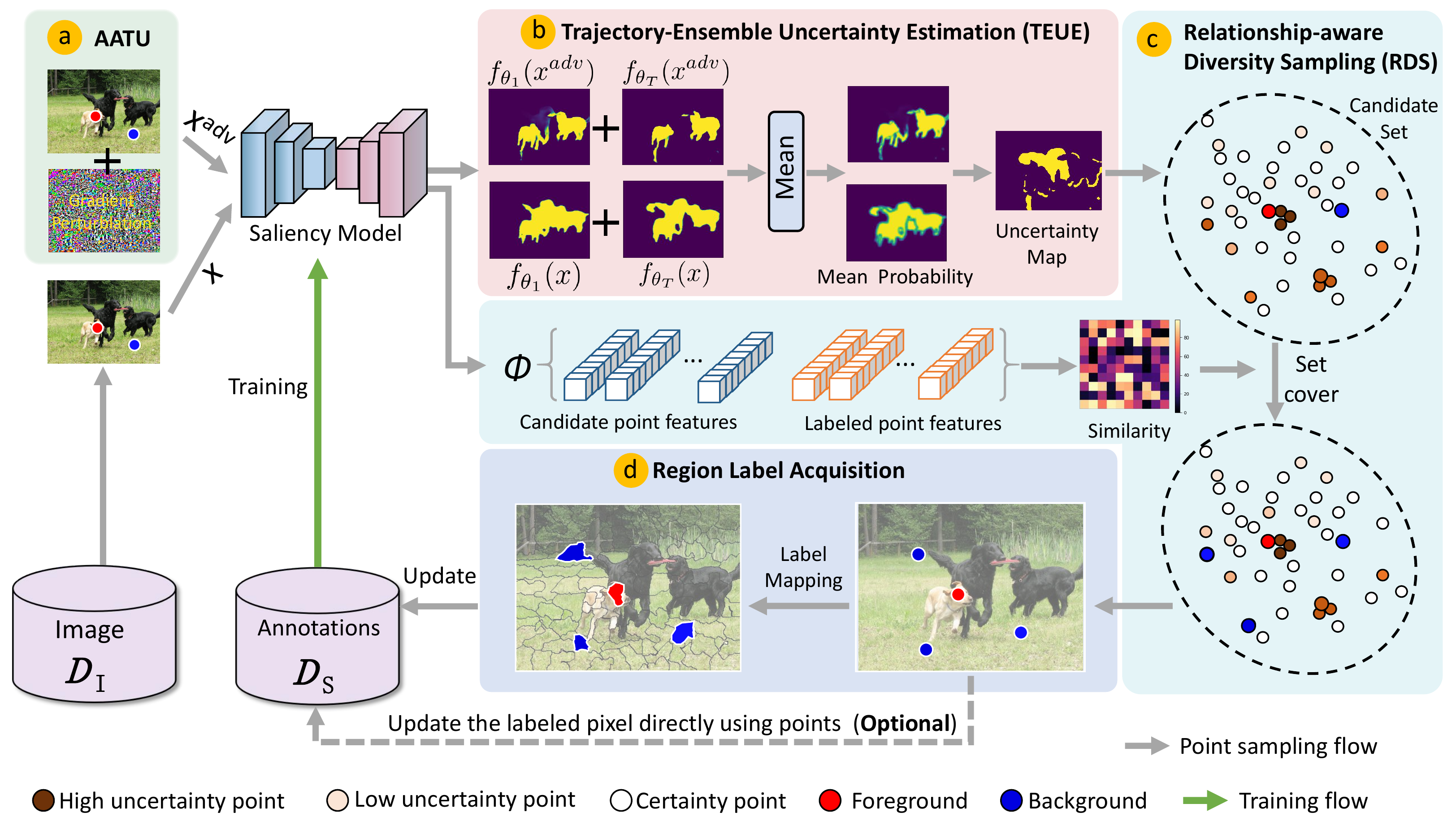}\vspace{-1em}
\caption{\textbf{Overview of the ATAL.} \textbf{First}, we train a saliency network on the initial labeled clean data $x$ and employ the PGD to generate the adversarial image $x^{adv}$. \textbf{Second}, we use the proposed TEUE to calculate the uncertainty score for clean data $x$ and adversarial data $x^{adv}$, and then obtain the candidate set. \textbf{Third}, we select a batch of uncertain pixels from the candidate set by using our RDS. \textbf{Finally}, we propagate the annotated pixel label to its corresponding superpixel block and update the current labeled set ${D}_S$ for the next round of training. This process is repeated until reaching the largest budget or desired performance.}
\label{fig:pipeline}
\end{figure*}

\section{Methodology}

The proposed ATAL framework (see Fig. \ref{fig:pipeline}) consists of four modules: \textbf{1)} a surprisingly effective adversarial attack for better identify uncertain pixels,  \textbf{2)} a lightening trajectory-ensemble networks for uncertainty estimation, \textbf{3)} a novel relationship-aware diversity sampling strategy for pixel sampling, and \textbf{4)} an effective region label acquisition method.

\subsection{Adversarial Attack Triggering Uncertainty}

The proposed AATU  explicitly identifies these uncertain pixels via injecting the adversarial perturbation into the input image, providing a new insight for uncertainty estimation. 
Concretely, given a saliency network ${f}$ and an input $x$, the output $o = {f}(x)$, where $x \in \mathbb{R}^{H \times W \times 3}$ and $o \in \mathbb{R}^{H \times W}$. 
For a clean sample $x$, pixel $x(i,j)$ is called clean pixel; for the adversarial sample $x^{adv}$, which is obtained by injecting adversarial perturbation into $x$, pixel $x^{adv}(i,j)$ is called adversarial pixel. In practice, we employ the standard first-order adversarial attack, i.e., Projected Gradient Descent (PGD) \cite{kurakin2016adversarial}, which iteratively updates the adversarial example under the $l_{\infty}$-norm threat model by:
\begin{equation}
x^{adv_{t+1}} = \operatorname{Clip} ( x^{adv_t} + \alpha \cdot \operatorname{sign}( \nabla_{x^{adv_t}} ( \mathcal{L} ( {f}(x^{adv_t}), y )) ) )
\end{equation}
where $x^{adv_t}$ is the adversarial sample after the $t$-th attack step, $\operatorname{Clip}(\cdot)$ forces it ouput to reside in the range of $[x - \epsilon, x + \epsilon]$, $\epsilon$ is the perturbation range, $\operatorname{sign}(\cdot)$ is the sign function, and $\alpha$ is the step size.

We further propose to divide these adversarial pixels into different categories according to their anti-perturbation ability. As shown in Fig. \ref{fig:clean_adv}, adversarial pixels can be classified into three categories: Safety Region (SR), Perturbation Insensitive Region (PIR), and Perturbation Sensitive Region (PSR).
\textbf{1) SR:} clean pixels and their corresponding adversarial pixels are classified into the same category in the output space. These safety pixels are robust to adversarial attack and commonly stay far away from the decision boundary. 
\textbf{2) PIR:} clean pixels and their corresponding adversarial pixels are classified into different categories. These clean pixels stay far away from the decision boundary, while adversarial pixels stay near the decision boundary. If we enhance the model's robustness, the PIR can be turned into SR.
\textbf{3) PSR:} clean pixels and their corresponding adversarial pixels are classified into different categories, and these clean pixels stay near the decision boundary.

According to the definition, it is clear that these most certain pixels are located in SR while these most uncertain pixels are located in PSR.
By injecting the PGD attack, we can easily identify these certain pixels that $x^{clean}(i,j)$ and $x^{adv}(i,j)$ have the same prediction. However, capturing these most uncertain pixels is non-trivial because they are entangled with these perturbation-insensitive pixels. We employ the DEN to distinguish these most uncertain pixels from PIR by enhancing the robustness of PIR and calculating the uncertainty score for each pixel. Despite its outstanding performance (see the 2nd row of Table \ref{tab:DEN}), the DEN introduces prohibitive training costs, which will be addressed in the following section.

\noindent\textbf{Advantages:} We surprisingly find that these perturbation-sensitive pixels triggered by adversarial attacks correspond to these most uncertain pixels. Moreover, our AATU can alleviate the overconfidence of existing AL methods (see Fig. \ref{fig:overconfidence}) and accurately locate these uncertain pixels.

\begin{figure*}[!h]
\centering
\resizebox{0.95\textwidth}{!}{
\begin{subfigure}{.3\textwidth}
\centering
\includegraphics[width=\textwidth]{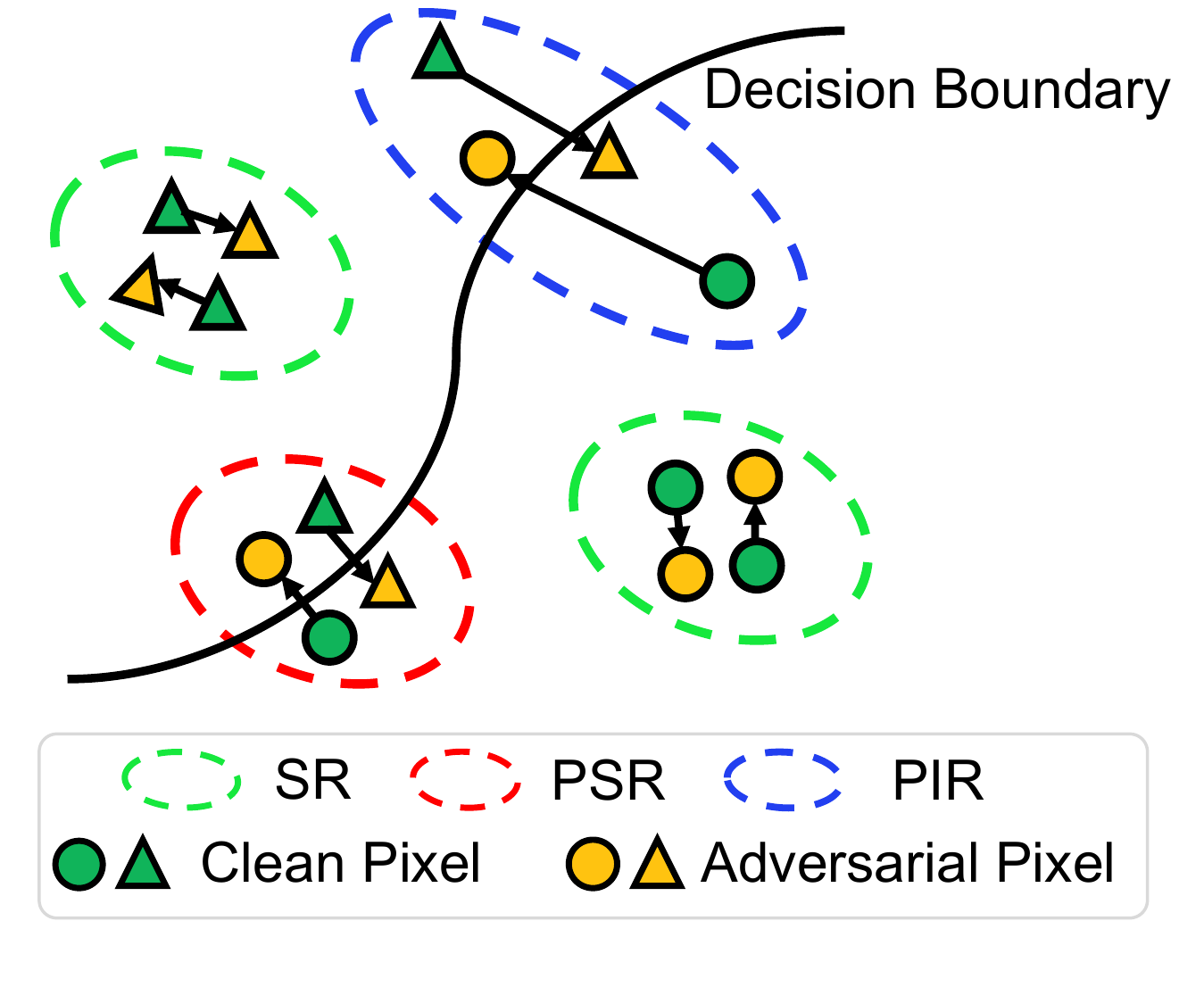}\vspace{-0.80em}
\caption{}
\label{fig:clean_adv}
\end{subfigure}
\begin{subfigure}{.33\textwidth}
\centering
\includegraphics[width=\textwidth]{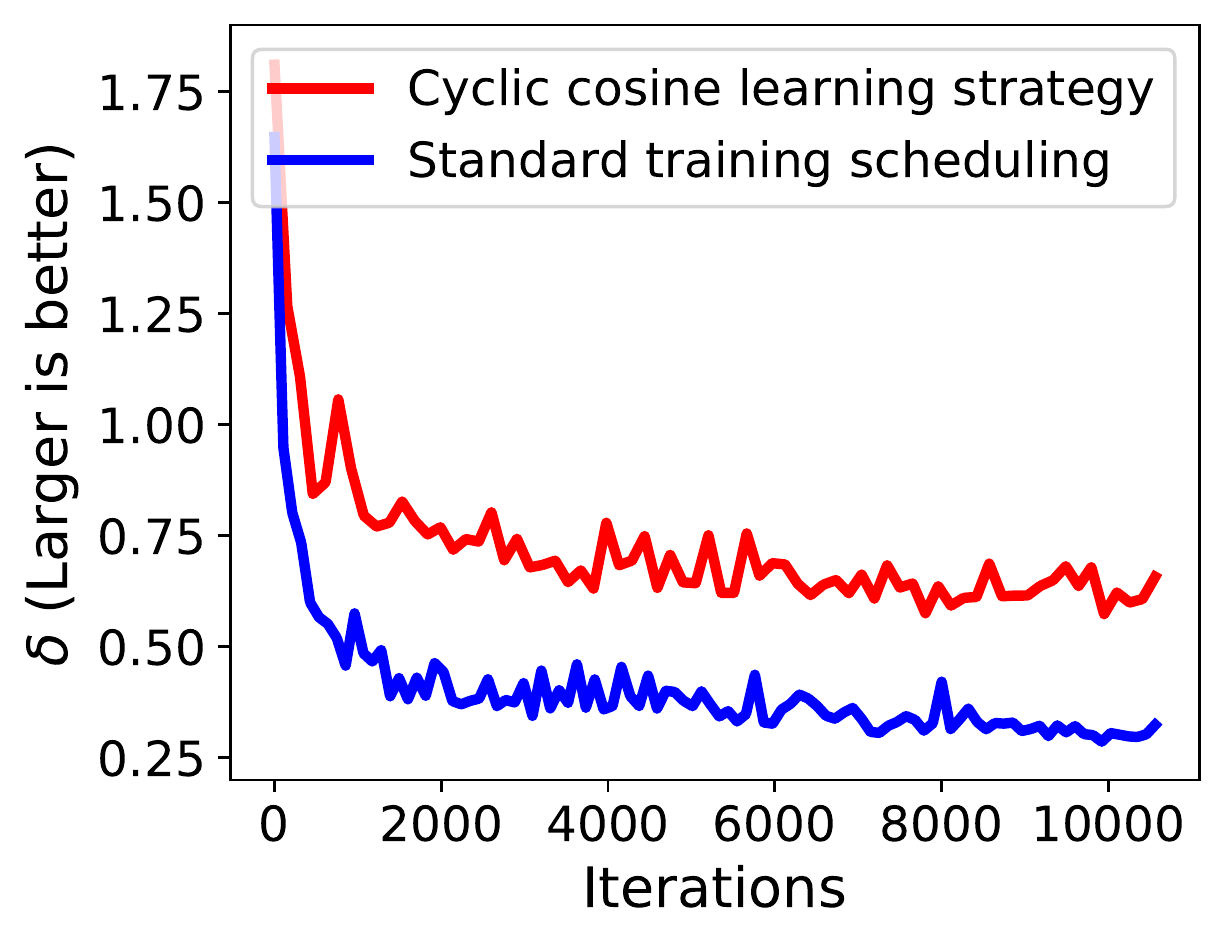}\vspace{-0.80em}
\caption{}
\label{fig:homogenization}
\end{subfigure}
\begin{subfigure}{.33\textwidth}
\centering
\includegraphics[width=\textwidth]{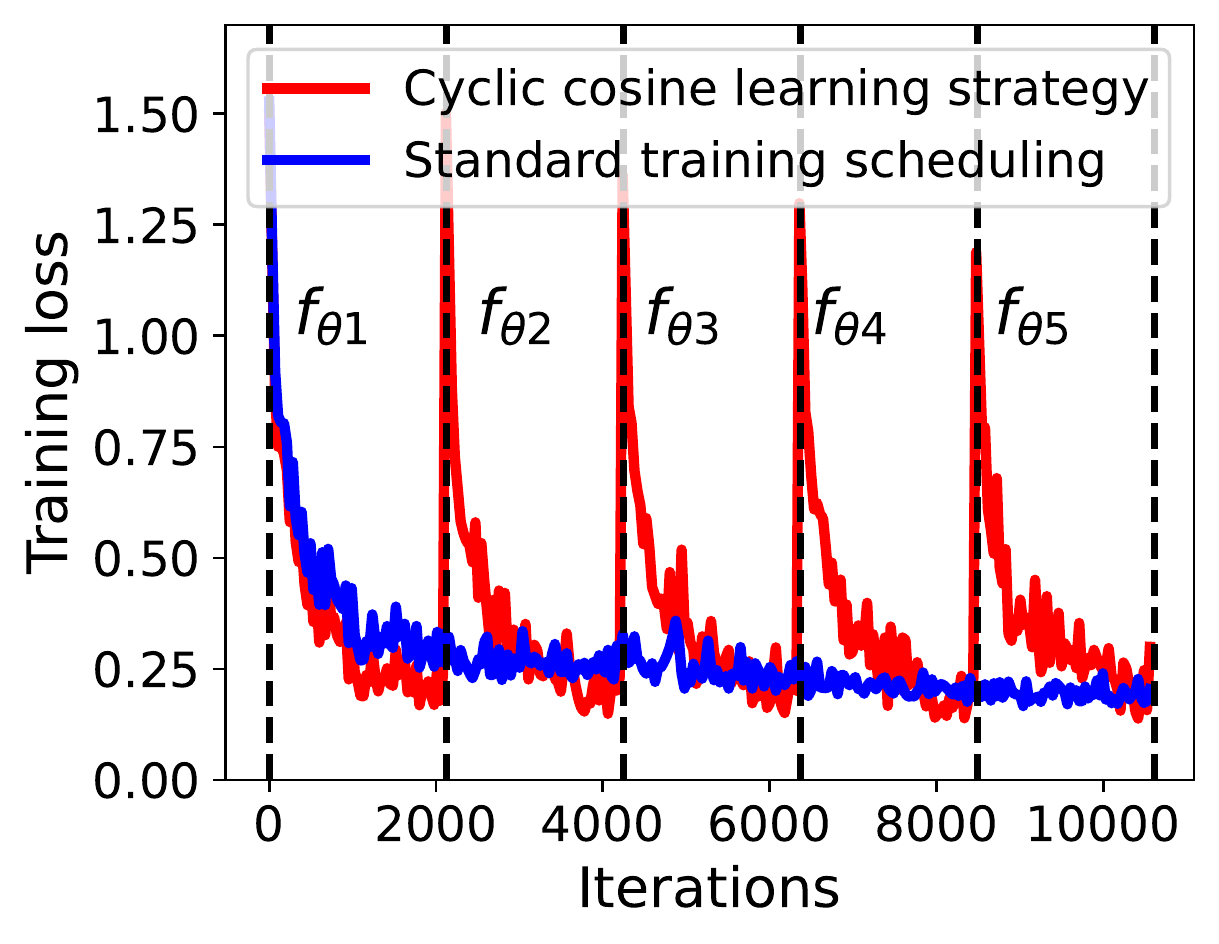}\vspace{-0.80em}
\caption{}
\label{fig:cca}
\end{subfigure}}\vspace{-1em}
\caption{(a) The illustration of the SR, PIR, and PSR division. (b) Our CCLS can alleviate the homogenization phenomenon of trajectory-ensemble networks. (c) Our CCLS can force the model to converge to its local minimum just a few epochs, which ensure that our TEUE can work like deep ensemble networks. }
\end{figure*}

\subsection{Trajectory-Ensemble Uncertainty Estimation}
Our TEUE mainly focuses on reducing the computational overheads stemming from repeating network training while attempting to retain the benefits brought by deep ensemble networks (DEN).
We begin by revisiting the procedure of the traditional ensemble method for uncertainty estimation. 
\textbf{First}, a set of $\{f_1, f_2,..., f_N\}$  train on the current labeled dataset $\{x_i,y_i\}_{i=1}^{M}$ with different initializations, forming $N$ well-trained model with different predictions. 
\textbf{Second}, these clean images $x$ and their adversarial counterparts $x^{adv}$ are then fed into these networks to obtain the probability maps $p(x)$ and $p(x^{adv})$:
\begin{equation}\vspace{-0.5em}
p(x)=\frac{1}{N}\sum_{i=1}^{N} f_i(x),\quad  p(x^{adv})=\frac{1}{N}\sum_{i=1}^{N} f_i(x^{adv})
\label{eq:deepensemble}
\end{equation}
\textbf{Finally}, the obtained  $p(x)$ and $p(x^{adv})$ can be used to calculate the uncertainty score map $u(x)$:
\begin{equation}\vspace{-0.5em}
u(x) = \operatorname{Diff}\left(\operatorname{BvSB}\left(u\left(x\right)\right), \operatorname{BvSB}\left(u\left(x^{adv}\right)\right)\right) \\
\label{eq:2}
\end{equation}
where $\operatorname{BvSB}(\cdot)$ is the best-versus-second best margin \cite{joshi2009multi},  $ \operatorname{Diff}(x_1, x_2)$ denotes these pixels where $x_1$ are inconsistency with $x_2$.

\begin{table}[t]
\center{
\Large{
\renewcommand\arraystretch{1}
\resizebox{0.48\textwidth}{!}{
\begin{tabular}{c||c||cc|cc}
\Xhline{1pt}
\multirow{2}{*}{}
&\multirow{1}{*}{Training}
& \multicolumn{2}{c|}{DUT-OMRON}
& \multicolumn{2}{c}{DUTS-TE}\\
\cline{3-6}
& cost&max$F_\beta\uparrow$  & S-m$\uparrow$    &max$F_\beta\uparrow$  & S-m$\uparrow$   \\ \hline\hline

Vanilla trainning& M & \textcolor{gray}{.7254}  &  \textcolor{gray}{.7367}   & \textcolor{gray}{.7852}  & \textcolor{gray}{.7768}  \\
DEN & MN             &.7716  &  .7930   & .8383  & .8325  \\
TUTE w/o CCLS & M    & \textcolor{gray}{.7336}  &  \textcolor{gray}{.7483}   & \textcolor{gray}{.7964}  & \textcolor{gray}{.7875}  \\
TUTE w/ CCLS & M    & {.7708} & {.7921}&  {.8374} & {.8326}     \\

\Xhline{1pt}
 \end{tabular}  }}
\vspace{-0.5em}
\caption{ Our DEN can achieve similar performance to DEN while reducing the computational cost to $1/N$.}
\label{tab:DEN}
}
\end{table}

Let’s consider the training cost of one epoch. We denote the cost of a single forward and backward pass for one image as 1, and the dataset size as M. Then, the cost of vanilla training for one epoch is:
\begin{equation}\vspace{-0.5em}
\operatorname{Cost}(\operatorname{vanilla}) = M \\
\label{eq:3}
\end{equation}
Similarly, the training cost of DEN is:
\begin{equation}\vspace{-0.5em}
\operatorname{Cost}(\operatorname{DEN}) = M \times N \\
\label{eq:4}
\end{equation}
Obviously, DEN increases the training cost by a factor of $N$ compared to the vanilla training baseline.

To reduce the computational cost, we first give a naive attempt to simplify DEN by ensembling the pre-recorded weights of model optimization trajectories, called trajectory-ensemble. Thus, we reformulate the Eq. \ref{eq:deepensemble} as:
\begin{equation}\vspace{-0.5em}
p(x) =  \frac{1}{T} \sum_{t=1}^{T} f_{\theta_t}(x), \quad p(x^{adv}) =  \frac{1}{T} \sum_{t=1}^{T} f_{\theta_t}(x^{adv})
\label{eq:5}
\end{equation}
where $f_{\theta_t}$ is the $t$-th snapshot of model optimization trajectory. As shown in the 3rd row of Table \ref{tab:DEN}, this strategy severely degrades the original DEN’s performance (i.e., $73.4\%$ vs.$ 77.2\%$ on the DUT-OMRON dataset).

After carefully diagnosing, we observed that, in the trajectory-ensemble setting, the model $f_{\theta_t}$ obtained at relatively late stages loses the superiority of the ensemble due to the ``homogenization phenomenon'' as shown in Fig. \ref{fig:homogenization} (blue line). We define the homogenization of a model by calculating the difference of these output of trajectory models over a period of iteration $\tau$:
\begin{equation}\vspace{-0.5em}
\delta =  \frac{1}{|\tau|}\sum_{t}^{t+\tau} |f_{\theta_{t+1}}(x) - f_{\theta_{t}}(x)|
\label{eq:2}
\end{equation}

We introduce a novel cyclic cosine learning strategy (CCLS) to address the homogenization of trajectory-ensemble networks. Concretely, we start training the model with a very large learning rate and then lower it at a very fast pace, forcing the model to converge to its local minimum after just a few epochs. The optimization is then restarting at a larger learning rate, which perturbs the model and moves it away from the local minimum. We repeat this procedure several times to obtain multiple networks. Formally, our cyclic cosine learning rate is defined as:
\begin{equation}\vspace{-0.5em}
\eta = \frac{1}{2}(\eta_{max} - \eta_{min})\left(1 + \operatorname{cos}\left( \pi\frac{ \operatorname{mod}\left(i-1, [I/L]\right)}{[I/L]} \right)\right)
\label{eq:2}
\end{equation}
where $\eta_{max}$ and $\eta_{min}$ are ranges for the learning rate, $i$ is the iteration number, $I$ is the total number of training iterations, and $L$ is the number of training cyclic. Fig. \ref{fig:cca} shows the difference between the standard training scheduling and our CCLS.  After training $L$ cycles, we obtained $L$ model snapshots $\{f_1,..., f_L\}$, each of which will be used in the final ensemble. As shown in Fig. \ref{fig:homogenization}, our CCLS can relieve the homogenization phenomenon.  

\noindent\textbf{Advantages:} As shown in the 4th row of Table \ref{tab:DEN}, our TEUE not only largely reduces the training cost to $1/N$ but also maintains the advantages brought by the ensemble networks.

\subsection{Relationship-aware Diversity Sampling}
\label{diversity_sampling}
Our proposed relationship-aware diversity sampling (RDS) aims to conquer the oversampling issue (selected points tend to be distributed in the most highlight region, see the uncertainty maps of Fig. \ref{fig:our_vs_others}) by considering the relationship among these labeled pixels.
Given an uncertainty map $u(x)$, unlike previous methods simply selecting the top-$k$ uncertainty points, we first sample the top $K\%$  pixels forming the candidate set $D_C$. We then select $k$ points from the candidate set $D_C$, where the selected points should satisfy the following properties: \textbf{1)} covering the candidate set $D_C$; \textbf{2)} dissimilar with the current labeled set $D_S$.

To this end, we first need to find a set of points $V$ that cover the candidate set $D_C$ as possible, which can be viewed as a variant of the set cover problem \cite{feige1998threshold}. 
Let $\sigma(p_u, p_v)$ denote the  Euclidean distance between a pair of points $p_u$, $p_v \in D_C$.
Our objective is to find a set  $V \subseteq D_C $ to minimize the maximum distance of any point of $D_C$ to its closest cluster center, which is an NP-hard problem.

\begin{figure}[t]
\centering
\resizebox{0.48\textwidth}{!}{
\includegraphics[width=1\textwidth]{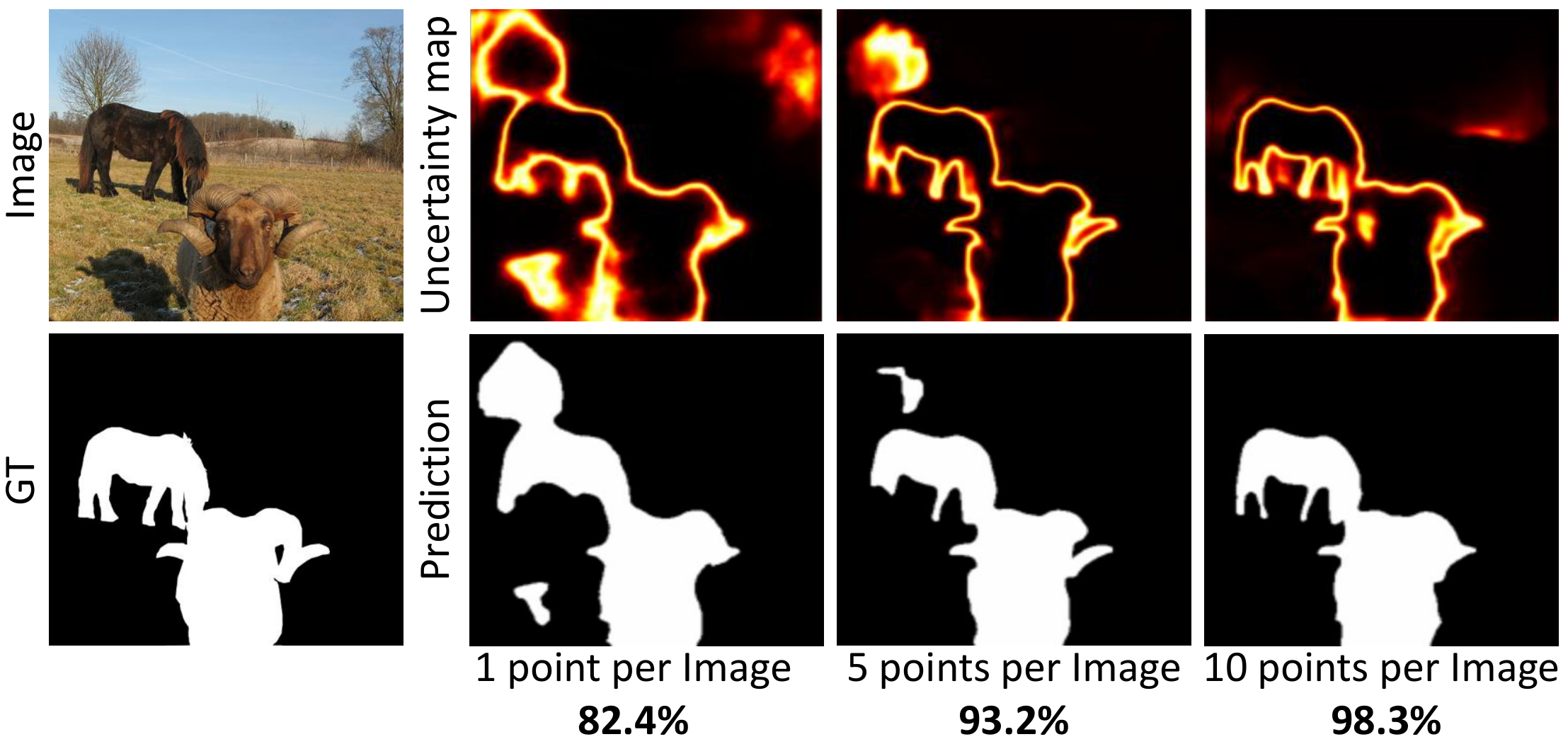}
}\vspace{-0.8em}
\caption{Our ATAL achieves $98.3\%$ performance of its fully-supervised version with $10$ labeled pixels per image.}
\label{fig:our_vs_others}
\end{figure}

\begin{equation}
\label{eq:min_max}
\min_{V \subseteq D_C } \sum_{p_u \in D_C} \sum_{p_v, \in V} \sigma(p_u, p_v).
\end{equation}
Instead of iterating through each pixel of the candidate set $D_C$, we introduce a greedy approximation algorithm (see Algorithm \ref{alg:Alg1 }) to find pixels $p_u \in D_C$ that maximizes $\sum_{p_v \in V}\sigma(p_u, p_v)$.

After obtaining the cover set $V$, we then  calculate the similarity of $D_S$ for candidate point $p_x \in V$ as below:

\begin{equation}
\phi(p_x, D_S)=  \frac{1}{|D_S|}\sum_{p_j \in D_S} \operatorname{Sim}(p_x, p_j),
\end{equation}
where $\operatorname{Sim}(\cdot)$ is commonly used $cosine$ similarity, $|D_S|$ denotes the number of labelled points. Finally, we rank all pixels in $V$ using the $\phi(p_x, D_S)$ and select the top-$k$ dissimilar points.  In this way, the selected points not only have a high uncertainty score but also consider the relationship between the candidate points and the labeled points.

\noindent\textbf{Advantages:} Our proposed RDS can address the oversampling issues by considering the relationship of these labeled pixels, achieving high performance with a few labeled pixels (see Fig. \ref{fig:our_vs_others}).




\begin{algorithm}
\caption{Our greedy approximation algorithm.}
\label{alg:Alg1 }
\begin{algorithmic}[1]
\REQUIRE a candidate set $D_C$, an integer $m \in \mathbb{N}$
\ENSURE $V \subseteq D_C $, $|V| = m$, with minimum max distance to points of $D_C$ as eq. \ref{eq:min_max}, 
$V \leftarrow p_u$, for $p_u \in D_C$ arbitrary

\WHILE {$|V| < m$}
\STATE let $p_u \in D_C \setminus V$ be the element maximizing $\sum_{p_v \in V}\sigma(p_u, p_v)$
\STATE update the $V$, $V \leftarrow p_u$
\ENDWHILE 
\RETURN $V$
\end{algorithmic}
\end{algorithm}

\subsection{Region Label Acquisition}
To exploit the structure and locality of the image, we argue that selecting regions instead of single pixels can achieve better performance.
Unlike previous region-based AL methods, which select the closest rectangle, we propose to map the annotated pixel label to its superpixel block. In this work, we employ an off-the-shelf SEEDS algorithm \cite{van2012seeds} due to its good performance in ensuring class coherency within each superpixel. Please note that our proposed framework is universal, and any other superpixels algorithms can also be employed.
\begin{table*}[t]
\centering
{
\renewcommand\arraystretch{0.8}
\resizebox{0.98\textwidth}{!}{
\begin{tabular}{c|r||cccccccc||cccccccc}
\Xhline{1pt} \rule{0pt}{11pt}            &        & \multicolumn{8}{c||}{\textbf{Fully-Supervised Models}}                                                      & \multicolumn{8}{c}{\textbf{Semi/Weakly-Supervised Models}}                              \\ \cline{3-18}
           & Metric & DGRL    & PAGR    & BAS     & CPD      & SAMN & MINet   & F3Net                     & PFSN           & MWS     & ENDS           & WS$^3$A    & MFNet           & FCS    & {\textbf{MINet$_{10}^{*}$}}   & {\textbf{F3Net$_{10}^{*}$}} & {\textbf{PFSN$_{10}^{*}$}}           \\\hline\hline
\multirow{6}{*}{\rotatebox{90}{{DUTS-O}}}
           & maxF$\uparrow $  & .7742   & .7709   & .8053   & .7966    & .8026 & .8098   & \textbf{\textcolor{blue}{.8133}}        & \textbf{\textcolor{red}{.8233}}  & .7176   & .7581          & .7532   & .6874          & .7170   &.7895 & \textbf{\textcolor{blue}{.7908}} &  \textbf{\textcolor{red}{.8030}} \\

           & S-m$\uparrow $   & .8059   & .7751   & .8362   & .8248     & .8299  & .8329  & \textbf{\textcolor{blue}{.8385}}                  & \textbf{\textcolor{red}{.8425}}  & .7559   & .7832          & .7848   & .7258          & .7448  & .8126 & \textbf{\textcolor{blue}{.8131}} & \textbf{\textcolor{red}{.8252}} \\

           & MAE$\downarrow$    & .0618   & .0709   & .0565   & .0560   & .0652  & .0555   & \textbf{\textcolor{red}{.0526}}          & .0545           & .1086   & .0759        & .0684  & {.0982}         & .0656 & .6130 & \textbf{\textcolor{blue}{.0604}} & \textbf{\textcolor{red}{.0584}} \\
         
           & avgF$\uparrow$   & .7656   & .7354   & .7875   & .7770    & .7655   & .7907  & \textbf{\textcolor{blue}{.7957}}                    & \textbf{\textcolor{red}{.8069}}  & .6777   & .7246          & .7386   &   .6608        & .7073 &.7764  & \textbf{\textcolor{blue}{.7834}} & \textbf{\textcolor{red}{.7862}} \\ 

           & W-F$\uparrow$   & .7093   & .6012 & \textbf{\textcolor{red}{.7520}} & .7047  & .6483 & .7194  & .7200  & \textbf{\textcolor{blue}{.7422}} & .4230 & .5372 &  .6518 &  .5104 & .6205 &.7014 & \textbf{\textcolor{blue}{.7109}} &\textbf{\textcolor{red}{.7133}} \\

           & E-m$\uparrow$   & .8425   & .6036 & \textbf{\textcolor{red}{.8573}} & .7874  & .6882 & .7975 & .8031  &  \textbf{\textcolor{blue}{.8134}}& .3364 &  .6161 & .7854 &  .6005& .7257 & .7761 & \textbf{\textcolor{red}{.8129}} & \textbf{\textcolor{blue}{.8027}} \\ \Xhline{0.6pt}

\multirow{6}{*}{\rotatebox{90}{{DUTS-TE}}}
           & maxF$\uparrow$   & .8287   & .8545   & .8591   & .8654     & .8360  & .8835  & \textbf{\textcolor{blue}{.8905}}                   & \textbf{\textcolor{red}{.8949}}  & .7686   & .8173          & .7889   & .7632          & .8296 &.8632  & \textbf{\textcolor{blue}{.8624}}& \textbf{\textcolor{red}{.8716}} \\

           & S-m$\uparrow$    & .8410   & .8369   & .8649   & .8684    & .8479 & .8834  & \textbf{\textcolor{blue}{.8881}}                    & \textbf{\textcolor{red}{.8916}}  & .7573   & .8190          & .8021   & .7764          & .8206 & .8587 & \textbf{\textcolor{blue}{.8586}} & \textbf{\textcolor{red}{.8741}} \\

           & MAE$\downarrow$    & .0500   & .0562   & .0480   & .0438     & .0582  & .0375 & \textbf{\textcolor{red}{.0358}}           & \textbf{\textcolor{blue}{.0359}}           & .0920   & .0657        & .0628   & .0798          & .0459   & .0468 & \textbf{\textcolor{blue}{.0454}}&  \textbf{\textcolor{red}{.0433}} \\
           
           & avgF$\uparrow$     & .8209   & .8108   & .8261   & .8357    & .7920   & .8566  & \textbf{\textcolor{blue}{.8647}}                    & \textbf{\textcolor{red}{.8714}}  & .7311   & .7743        & .7715   &   .7310         & .8085   &.8357 & \textbf{\textcolor{blue}{.8381}} & \textbf{\textcolor{red}{.8491}}\\ 

           & W-F$\uparrow$     &  .7677  &  .6845 &  .7930 & .7689  &  .6816  & .7966 & \textbf{\textcolor{blue}{.8011}}  & \textbf{\textcolor{red}{.8198}} & .5312 & .6129 &  .6897 & .5948 & .7496 &.7704 & \textbf{\textcolor{blue}{.7748}}& \textbf{\textcolor{red}{.7996}} \\

           & E-m$\uparrow$     & \textbf{\textcolor{blue}{.8856}}  &  .6130  &  \textbf{\textcolor{red}{.8858}} &  .8377 &  .7124  & .8518 & .8640  & {.8737} &  .6103 & .6954 & .8214 &.6836  &   \textbf{\textcolor{blue}{.8616}}  &.8324 & .8548&  \textbf{\textcolor{red}{.8657}} \\ \Xhline{0.6pt}
           
\multirow{6}{*}{\rotatebox{90}{{ECSSD}}}
           & maxF$\uparrow$     & .9224   & .9268   & .9424   & .9392   & .9279  & .9445   & \textbf{\textcolor{blue}{.9453}}                    & \textbf{\textcolor{red}{.9523}}  & .8778   & .9002          & .8880   & .8727          & .9108   &.9236 & \textbf{\textcolor{blue}{.9290}}&  \textbf{\textcolor{red}{.9327}} \\

           & S-m$\uparrow$      & .9028   & .8892   & .9162   & .9181     & .9071 & .9239  & \textbf{\textcolor{blue}{.9242}}                    & \textbf{\textcolor{red}{.9298}}  & .8275   & .8707          & .8655   & .8366          & .8787   &  \textbf{\textcolor{blue}{.8957}} &{.8938} & \textbf{\textcolor{red}{.8993}} \\

           & MAE$\downarrow$      & .0407   & .0609   & .0370   & .0371    & .0501  & .0334  & \textbf{\textcolor{blue}{.0333}}                    & \textbf{\textcolor{red}{.0309}}  & .0963   & .0676        & .0590   & .0841          & .0471   &.0478 & \textbf{\textcolor{blue}{.0461}} & \textbf{\textcolor{red}{.0439}} \\
         
           & avgF$\uparrow$     & .9122   & .8944   & .8970   & .9216    & .8985  & .9295  & \textbf{\textcolor{blue}{.9272}}                    & \textbf{\textcolor{red}{.9346}}  & .8430   & .8730          & .8733   &  .8431         & .8951   &.9045 & \textbf{\textcolor{blue}{.9083}} & \textbf{\textcolor{red}{.9131}} \\ 

           & W-F$\uparrow$      & .8909  & .8218  & \textbf{\textcolor{blue}{.9043}}   &  .8889  & .8337 & .8880 & .9020  & \textbf{\textcolor{red}{.9142}} &  .6520 & .7862 & .8318 & .7508 &  .8678& .8752 & \textbf{\textcolor{blue}{.8773}}& \textbf{\textcolor{red}{.8807}} \\

           & E-m$\uparrow$      & \textbf{\textcolor{blue}{.9373}}  & .5582  & \textbf{\textcolor{red}{.9382}}  &  .9020 & .8010  &  .9012 &  .9150  & .9199 & .5554 & .8232 & .9033 & .7987  &  .9212  & .9124& \textbf{\textcolor{blue}{.9233}} &  \textbf{\textcolor{red}{.9335}} \\ \Xhline{0.6pt}
           
\multirow{6}{*}{\rotatebox{90}{{HKU-IS}}}
           & maxF$\uparrow$     & .9105   & .9176   & .9285   & .9251      & .9147 & .9351  & \textbf{\textcolor{blue}{.9368}}                   & \textbf{\textcolor{red}{.9428}}  & .8560   & .9041          & .8805   & .8746          & .8992   &.9112 & \textbf{\textcolor{blue}{.9149}}&  \textbf{\textcolor{red}{.9245}} \\

           & S-m$\uparrow$      & .8945   & .8873   & .9090   & .9055     & .8983 & .9160  & \textbf{\textcolor{blue}{.9173}}                    & \textbf{\textcolor{red}{.9244}}  & .8182   & .8838          & .8649   & .8523          & .8718   &.8905 &  \textbf{\textcolor{blue}{.8931}} & \textbf{\textcolor{red}{.8969}} \\

           & MAE$\downarrow$      & .0356   & .0475   & .0322   & .0342     & .0449  & .0285 & \textbf{\textcolor{blue}{.0280}}                    & \textbf{\textcolor{red}{.0259}}  & .0843   & .0461        & .0470   & .0582          & .0389   &.0364 & \textbf{\textcolor{blue}{.0359}}& \textbf{\textcolor{red}{.0346}} \\
          
           & avgF$\uparrow$     & .8968   & .8904   & .9046   & .9004     & .8856  & .9172 & \textbf{\textcolor{blue}{.9177}}                    & \textbf{\textcolor{red}{.9256}}  & .8291   & .8801          & .8677   &  .8478         & .8836   &.9062 & \textbf{\textcolor{blue}{.9071}} &  \textbf{\textcolor{red}{.9085}} \\ 

           & W-F$\uparrow$      & .8752   & .8049  & 8893 & .8660 & .8105  & .9026  & \textbf{\textcolor{blue}{.8900}}  & \textbf{\textcolor{red}{.9029}} & .6131  &  .7838 & .8250  &  .7492  & .8561  &\textbf{\textcolor{blue}{.8740}} & .8674 & \textbf{\textcolor{red}{.8761}}  \\

           & E-m$\uparrow$      & \textbf{\textcolor{red}{.9380}}   & .5065  & \textbf{\textcolor{blue}{.9356}} & .8879  & .7857 & .9059 & .9120  &  .9192 &  .5075 & .8164 & .9051  &   .8143 &   .9295  & .9376&  \textbf{\textcolor{red}{.9445}}& \textbf{\textcolor{blue}{.9398}}  \\ \Xhline{0.6pt}

\multirow{6}{*}{\rotatebox{90}{{PASCAL-S}}}
           & maxF$\uparrow$     & .8808   & .8691   & .8757   & .8841    & .8568  & .8894  & \textbf{\textcolor{blue}{.8948}}                    & \textbf{\textcolor{red}{.8986}}  & .8140   & .8706          & .8374   & .8230          & .8742   &.8762 &  \textbf{\textcolor{blue}{.8896}}&  \textbf{\textcolor{red}{.8916} }\\

           & S-m$\uparrow$      & .8278   & .7925   & .8194   & .8277   & .8027  & .8333 & \textbf{\textcolor{blue}{.8404}}                    & \textbf{\textcolor{red}{.8431}}  & .7532   & .8025          & .7805   & .7663          & .8102   &.8078 & \textbf{\textcolor{blue}{.8140}}& \textbf{\textcolor{red}{.8234} }\\

          & MAE$\downarrow$      & .0823   & .1149   & .0924   & .0890    & .1130   & .0828  & \textbf{\textcolor{blue}{.0799}}                  & \textbf{\textcolor{red}{.0790}}  & .1509   & .1144         & .1106   & .1310          & \textbf{\textcolor{red}{.0849}}   &.0923 &  .0915& \textbf{\textcolor{blue}{.0879}} \\
          
           & avgF$\uparrow$     & .8528   & .8148   & .8100   & .8439  & .8054   & .8512   & \textbf{\textcolor{blue}{.8580}}                   & \textbf{\textcolor{red}{.8614}}  & .7566   & .8222          & .8054   &   .7789         & .8387   &.8403 &  \textbf{\textcolor{red}{.8490}}& \textbf{\textcolor{blue}{.8445}} \\ 

           & W-F$\uparrow$      & .7806  &  .7008   & .7762   &  .7707  & .7239 &   .7789 & \textbf{\textcolor{blue}{.7890}}   & \textbf{\textcolor{red}{.7981}}  &  .6125 &  .7111  &  .7294 & .6704 &  .7704 & .7538&  \textbf{\textcolor{blue}{.7684}} & \textbf{\textcolor{red}{.7788}} \\

           & E-m$\uparrow$      & .8380   &  .5921   & .8340  &  .8273 & .7653  &   .8347  & \textbf{\textcolor{blue}{.8430}}   &  \textbf{\textcolor{red}{.8482}} & .5456 &  .8028  &  .8308 & .7533  &    .8560    &.8401 & \textbf{\textcolor{blue}{.8591}} & \textbf{\textcolor{red}{.8612}} \\ \Xhline{1pt}
           
\end{tabular}
}}\vspace{-0.7em}
\caption{Our ATAL can achieve about $97\%$ -- $99\%$ F-measure of its fully-supervised version with only 10 annotated pixels per image and outperform existing weakly-supervised methods by a large margin. The ``F3Net$_{10}^*$'' denotes F3Net trained on our ATAL selected point labeled datasets. \textbf{\textcolor{red}{Red}} and \textbf{\textcolor{blue}{Blue}} indicate the best and the second-best results.}
\label{tab:main_results}
\end{table*} 

\begin{table}[h]
\center
\renewcommand\arraystretch{1}
\resizebox{0.30\textwidth}{!}{
\begin{tabular}{l||lll}
\Xhline{1.2pt}
\diagbox{Train}{Test} & MINet & F3Net & PFSN \\ \hline \hline
MINet   &   \textbf{97.4\%}  &    96.4\%   &   96.1\%   \\
F3Net &  96.8\%   &   \textbf{97.2\%}    &   96.6\%   \\
PFSN  &  96.3\%   &    96.5\%   & \textbf{97.5\%}   \\
\Xhline{1.2pt}  
\end{tabular}
}\vspace{-0.5em}
\caption{Cross-validation results demonstrate that our ATAL has good generability to different SOD models.}
\label{tab:generability}
\end{table} \vspace{-0.5em}

\begin{figure*}[h]
\centering
\includegraphics[width=0.97\textwidth]{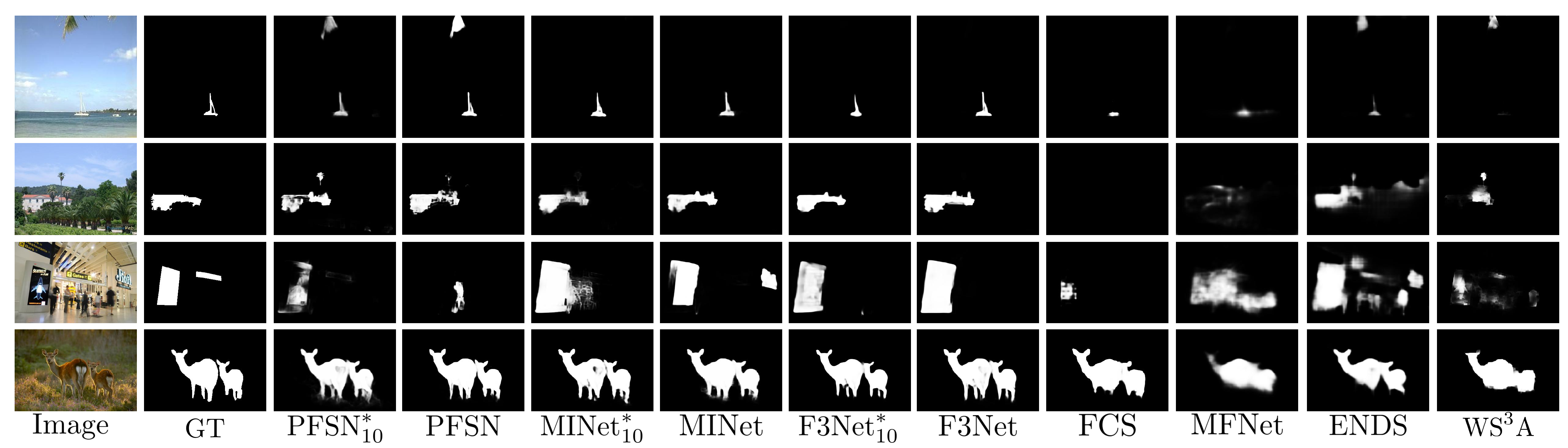}\vspace{-0.8em}
\caption{Our ATAL can identify an informative point-labeled dataset, where a saliency model trained on it can achieve the equivalent performance of its fully-supervised version.}
\label{fig:main_visual_comparisons}
\end{figure*}

\section{Experiments} 

\subsection{Experimental Setup}

\noindent{\textbf{Implementation Detail.}}
\textbf{Network.} In this work, we do not focus on network architecture design, in our experiments, we adopt F3Net \cite{wei2020f3net}, {MINet} \cite{pang2020multi} and PFSN~\cite{ma2021PFSNet} as our saliency model to validate the effectiveness of our ATAL algorithm. The detailed parameters setting of these saliency networks can be found in their paper.  
\textbf{Point Annotation.} For the point annotation collection, instead of using a real annotator, the point annotation process can be seamlessly combined with the DUTS-TR groundtruth. We can automatically classify the selected points into the correct category according to the positions in the groundtruth without any annotation efforts. 
\textbf{Loss Function.} In our ATAL algorithm, we treat each labeled pixel as an independent training sample. Thus, we can train these SOD models just like these fully-supervised models by minimizing the cross-entropy loss between prediction and annotated pixel in the same coordinate.  
We initially randomly select 2 points to label, and the max budget is set to 20. According to experimental results, when $K$=3, the saliency model obtains the best results. The number of training cyclic $L$  is set to 5.
\textit{Note that these adversarial images are not used to train saliency networks, and it only used in the point sampling procedure.}

\noindent{\textbf{Datasets and Evaluation Metrics.} }  We follow previous approaches to use the DUTS-TR \cite{wang2017learning} as the training set, and the only difference is that our model is supervised with point annotation. We evaluate our model on 5 datasets, including DUTS-OMRON \cite{yang2013saliency}, DUTS-TE \cite{wang2017learning}, ECSSD \cite{ecssd}, HKU-IS \cite{zhao2015saliency} and PASCAL-S \cite{li2014secrets}. We adopt several widely-used metrics to evaluate our method, including the {P}recision-{R}ecall, F-measure, Mean Absolute Error (MAE), S-measure~\cite{Smeasure}, and E-measure~\cite{fan2018enhanced}.

\subsection{Comparisons with State-of-the-Art (SOTA)}

We compare the our approach with 13 SOTA SOD models, including MWS \cite{zeng2019multi}, EDNS \cite{zhang2020learning}, WS$^3$A \cite{zhang2020weakly}, MFNet \cite{piao2021mfnet}, FCS \cite{zhang2021few}, {DGRL} \cite{DGRL18}, {PAGR}  \cite{PAGRN}, {BAS} \cite{BASNet19}, {CPD} \cite{CPD19}, {MINet} \cite{pang2020multi}, F3Net \cite{wei2020f3net}, PFSN~\cite{ma2021PFSNet}, and SAMN~\cite{liu2021samnet}.

\begin{figure*}[t]

   \centering
   \resizebox{1\textwidth}{!}{
   \begin{subfigure}{.21\textwidth}
     \centering
     \includegraphics[width=\textwidth]{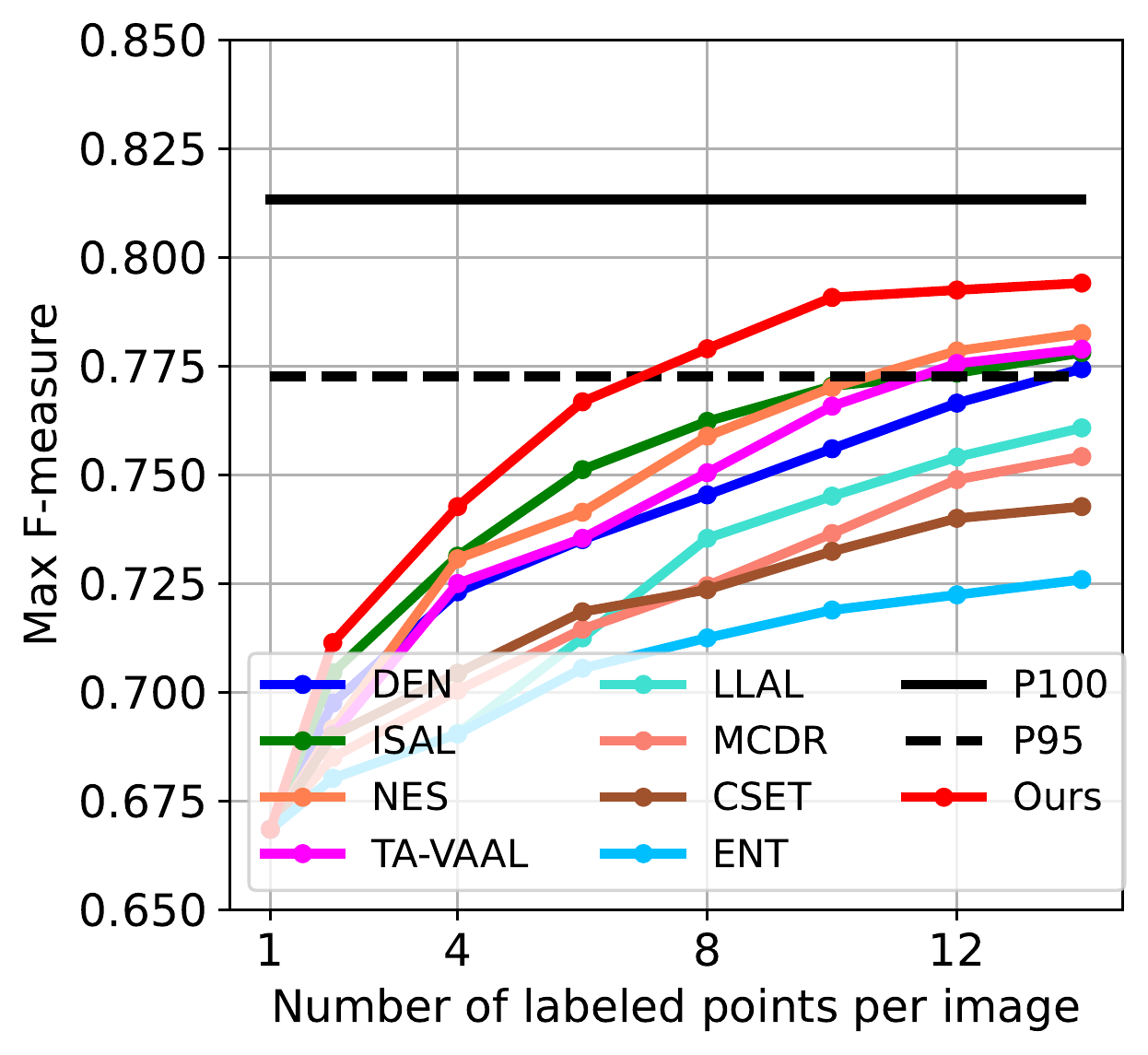}\vspace{-0.5em}
     \caption*{DUT-OMRON}
     \label{fig:pionts_per_image1}
   \end{subfigure}
   \begin{subfigure}{.21\textwidth}
     \centering
     \includegraphics[width=\textwidth]{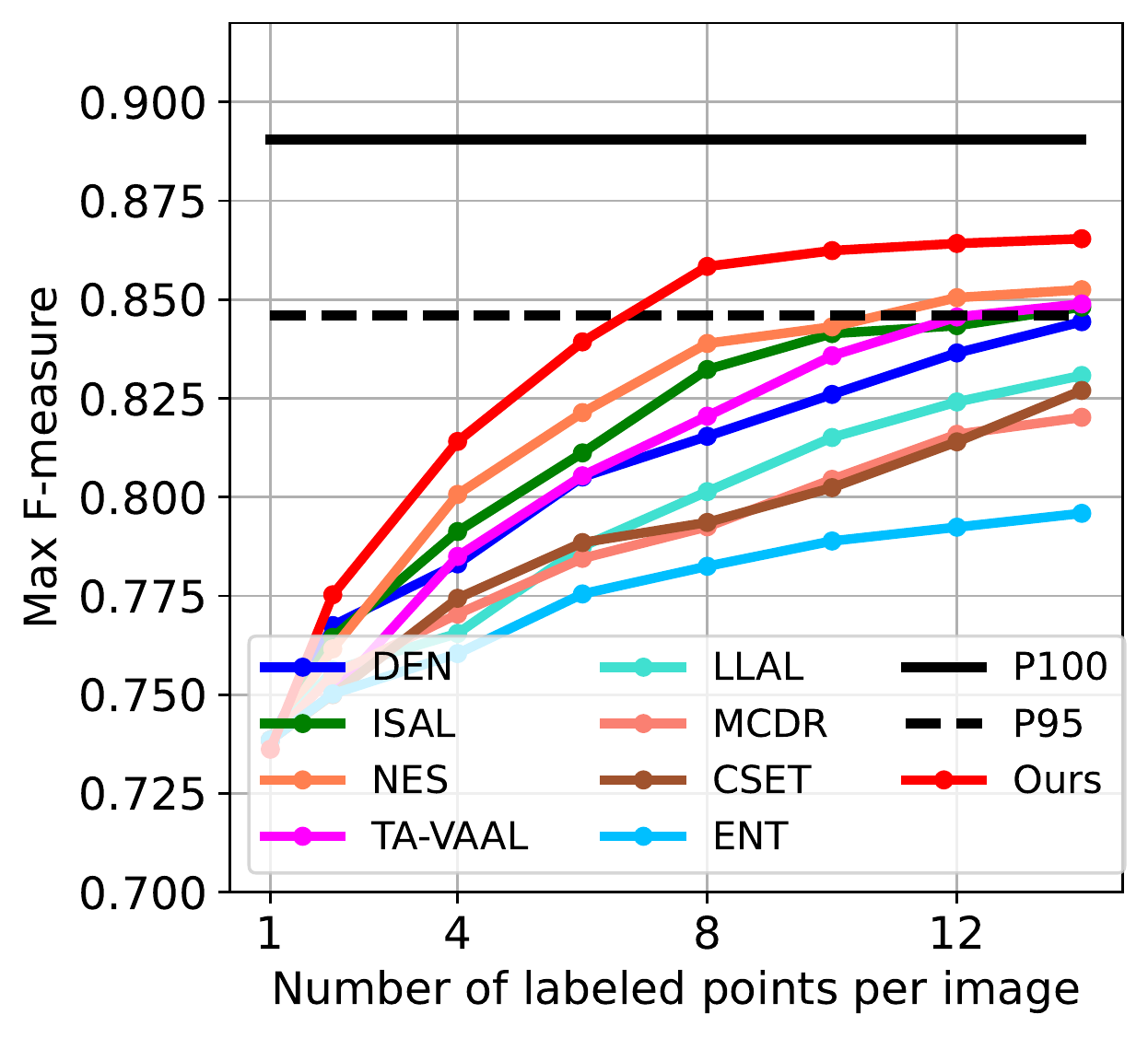}\vspace{-0.5em}
     \caption*{DUTS-TE}
   \end{subfigure}\hspace*{-0.38em}
   \begin{subfigure}{.205\textwidth}
     \centering
     \includegraphics[width=\textwidth]{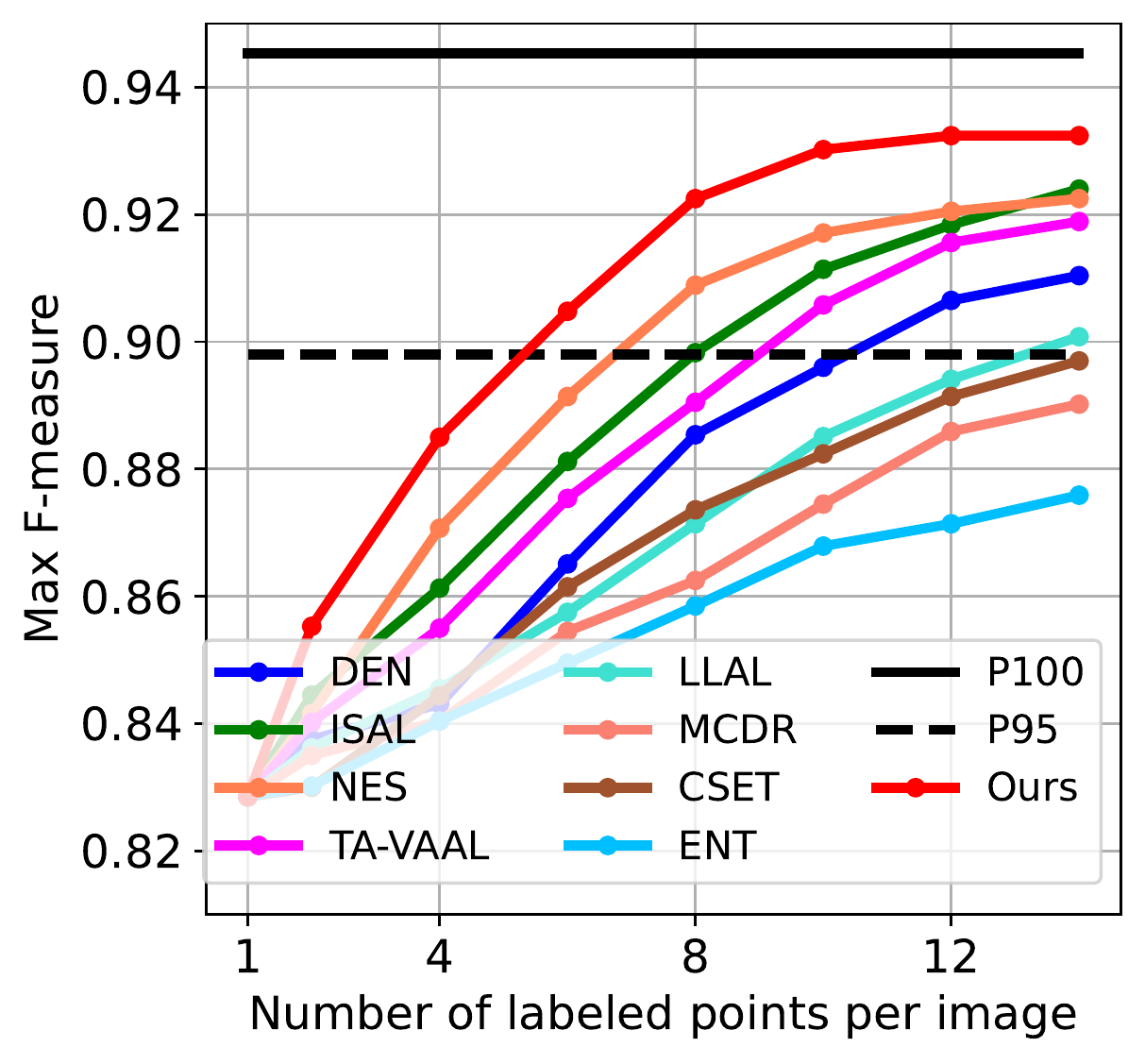}\vspace{-0.5em}
     \caption*{ECSSD}
   \end{subfigure}\hspace*{-0.38em}
   \begin{subfigure}{.21\textwidth}
     \centering
     \includegraphics[width=\textwidth]{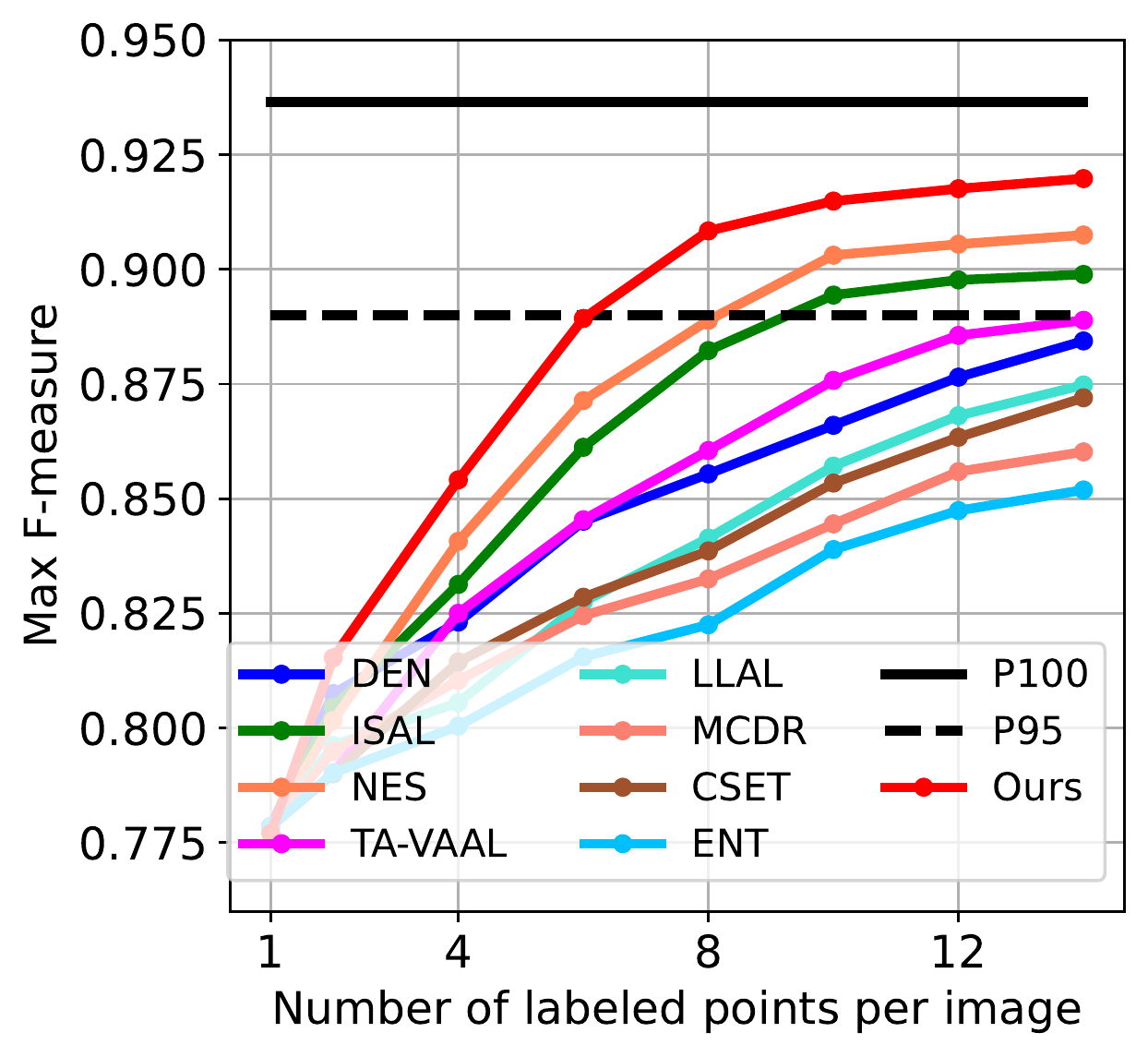}\vspace{-0.5em}
     \caption*{HKU-IS}
   \end{subfigure}
   \begin{subfigure}{.21\textwidth}
     \centering
     \includegraphics[width=\textwidth]{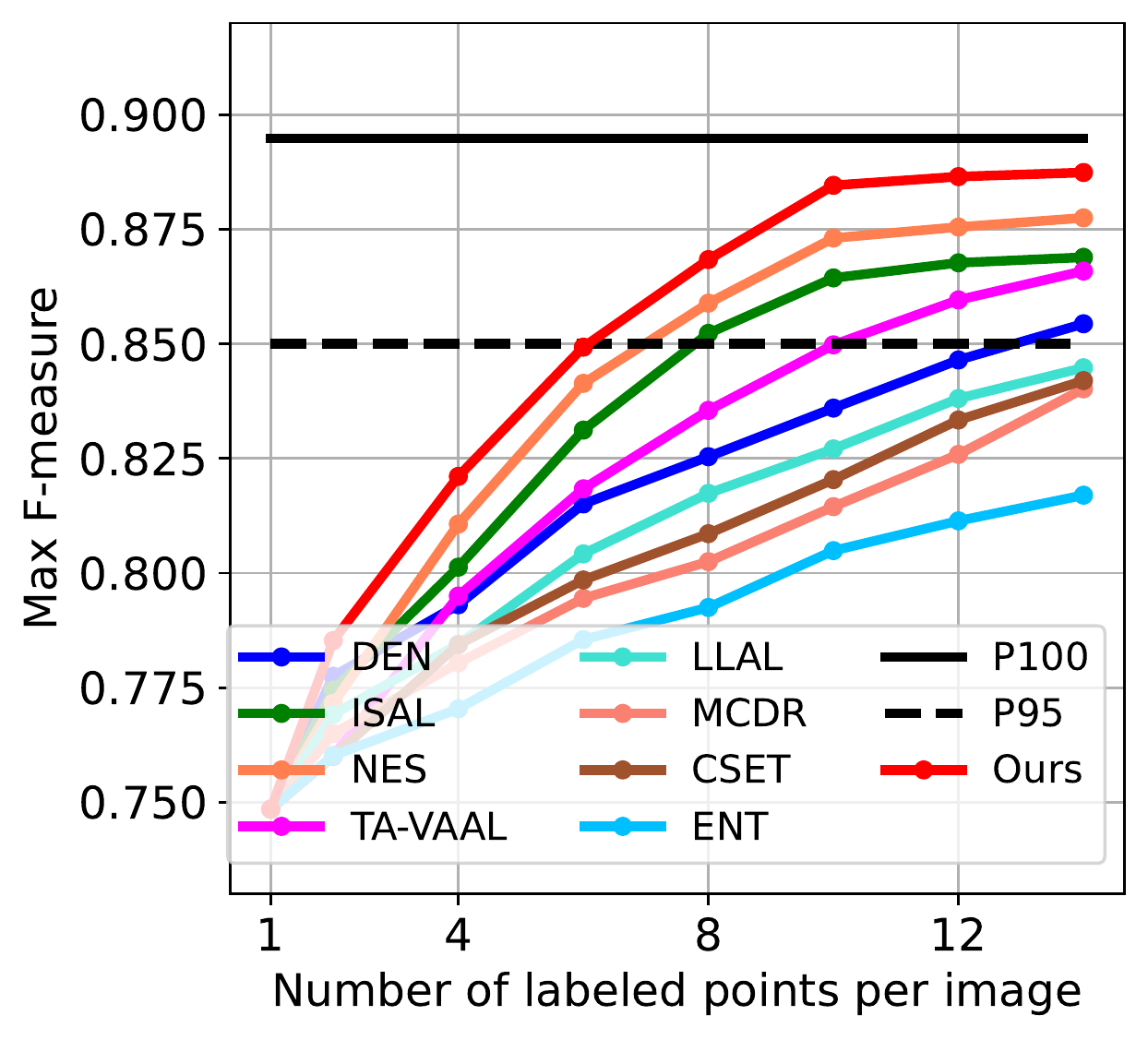}\vspace{-0.5em}
     \caption*{PASCAL-S}
   \end{subfigure}
     }
   \vspace{-2em}
   \caption{Our ATAL outperforms all other active learning methods and achieves $97\%$ -- $99\%$ of maximum model performance with just 10 labeled pixels per image. The horizontal solid line (p100) at the top represents model performance with the pixel-wise labeled training set. The dashed line (p95) represents $95\%$ of that performance.}
   \label{fig:our_vs_al}
\end{figure*}

\noindent\textbf{Quantitative Comparison.} 
To validate the effectiveness of our ATAL, we conduct experiments on three popular SOD models (i.e., MINet, F3Net, and PFSN). As shown in Table \ref{tab:main_results}, our ATAL can achieve about $97\%$ -- $99\%$ F-measure of its fully-supervised version with only ten annotated pixels per image and outperform existing weakly-supervised methods by a large margin.
Besides, we conducted nine cross-validations on MINet, F3Net, and PFSN. As shown in Table \ref{tab:generability}, the point dataset selected by F3Net can be used to train MINet and PFSN with negligible performance degradations. Similar observations can also be found in Minet and PFSN models. These results demonstrate the effectiveness of our ATAL, and also indicate our ATAL has good generability to different SOD models.

\noindent{\textbf{Qualitative Comparison.}} As shown in Fig. \ref{fig:main_visual_comparisons}, a saliency model trained on our ATAL selected point-labeled datasets can achieve the equivalent performance of its fully-supervised version. Moreover, our ATAL outperforms existing weakly-supervised SOD in various challenging scenarios, small objects (the 1st row), occlusion scenes (the 2nd row), cluttered backgrounds (the 3rd row), and low contrast (the 4th row).

\noindent{\textbf{Comparisons Against Active Learning Methods.}}  We compare our ATAL method with 8 commonly used AL methods, including \textbf{ISAL}\cite{liu2021influence}, \textbf{NES} \cite{zaidi2021neural}, \textbf{TA-VAAL}) \cite{kim2021task}, \textbf{LLAL} \cite{yoo2019learning}, \textbf{MCDR} \cite{mackowiak2018cereals}, \textbf{CSET} \cite{sener2017active}, \textbf{DEN} \cite{lakshminarayanan2017simple}, and \textbf{ENT} \cite{wang2016cost}.  As shown in Fig. \ref{fig:our_vs_al}, our ATAL achieves better performance than other AL methods with the same number of labeled data.

\subsection{Ablation Study of Our Innovation}

We conduct ablation study using the F3Net as saliency network, and take the deep ensemble networks (DEN) \cite{lakshminarayanan2017simple} as the baseline.

\noindent{\textbf{Number of Labeled Points Per Image.}} As shown in Fig. \ref{fig:pionts_per_image}(a), with the increase of labeled points, the performance of F3Net improves rapidly before ten labeled points per image. After that, the performance improvements were only about  $0.0044$ in max-F from 10 to 20. In our experiments, we use the 10-point-labeled dataset, which is 2$\times$ faster than the 20-point-labeled dataset.

\begin{table}[t]
\center{
\Large{
\renewcommand\arraystretch{1}
\resizebox{0.48\textwidth}{!}{
\begin{tabular}{c||ccccc||cc|cc}
\Xhline{1pt}
\multirow{2}{*}{No.}
&\multirow{2}{*}{DEN}
&\multirow{2}{*}{AATU}
& \multirow{2}{*}{TEUE}
& \multirow{2}{*}{RDS}
& \multirow{2}{*}{SP}

& \multicolumn{2}{c|}{DUT-OMRON}
& \multicolumn{2}{c}{DUTS-TE}
\\
\cline{7-10}
 && &  &  & &max$F_\beta\uparrow$  & S-m$\uparrow$    &max$F_\beta\uparrow$  & S-m$\uparrow$   \\
\hline\hline

\textcircled{1}& \CheckmarkBold & &&  & &.7562  &  .7682   & .8261  & .8126  \\
\textcircled{2}& \CheckmarkBold  &\CheckmarkBold& &  &  &.7716  &  .7930   & .8383  & .8325  \\

\textcircled{3}& &\CheckmarkBold  & \CheckmarkBold & & & {.7708} & {.7921}&  {.8374} & {.8326}     \\

\textcircled{4}& &\CheckmarkBold  & \CheckmarkBold & \CheckmarkBold & & {.7814} & {.8015}&  {.8479} & {.8421}    \\

\textcircled{5}& &\CheckmarkBold  & \CheckmarkBold & \CheckmarkBold &  \CheckmarkBold& \textbf{.7908} & \textbf{.8131}&  \textbf{.8624} & \textbf{.8586}    \\

\Xhline{1pt}
 \end{tabular}  }}
\vspace{-0.5em}
\caption{Ablation study on each component demonstrates that all components are necessary for the proposed ATAL.}
\label{tab:ablation_study}}
\end{table}

\noindent{\textbf{{Effectiveness of AATU.}}  We verify the effectiveness of AATU by combining our AATU with the model \textcircled{1}. As we can see, DEN+AATU can improve model \textcircled{1} performance by a considerable margin.

\noindent{\textbf{{Effectiveness of TEUE.}} To verify the effectiveness of TEUE, we further replace the DEN with our TEUE. As we can see, model \textcircled{3} can achieve similar performance to model \textcircled{2} while reducing the computational cost to $1/5$. 

\noindent{\textbf{Effectiveness of RDS.}} To verify the effectiveness of RDS, we replace the top-$k$ sampling strategy in model \textcircled{3} with our RDS ( model \textcircled{4}). As we can see, our RDS can further improve the performance of model \textcircled{3}. In Fig. \ref{fig:pionts_per_image}(b), we also show the distribution of labeled points on DUTS-TR, indicating the effect of our RDS.

\begin{figure}[t]

   \centering
   \resizebox{0.43\textwidth}{!}{
   \begin{subfigure}{.25\textwidth}
     \centering
     \includegraphics[width=\textwidth]{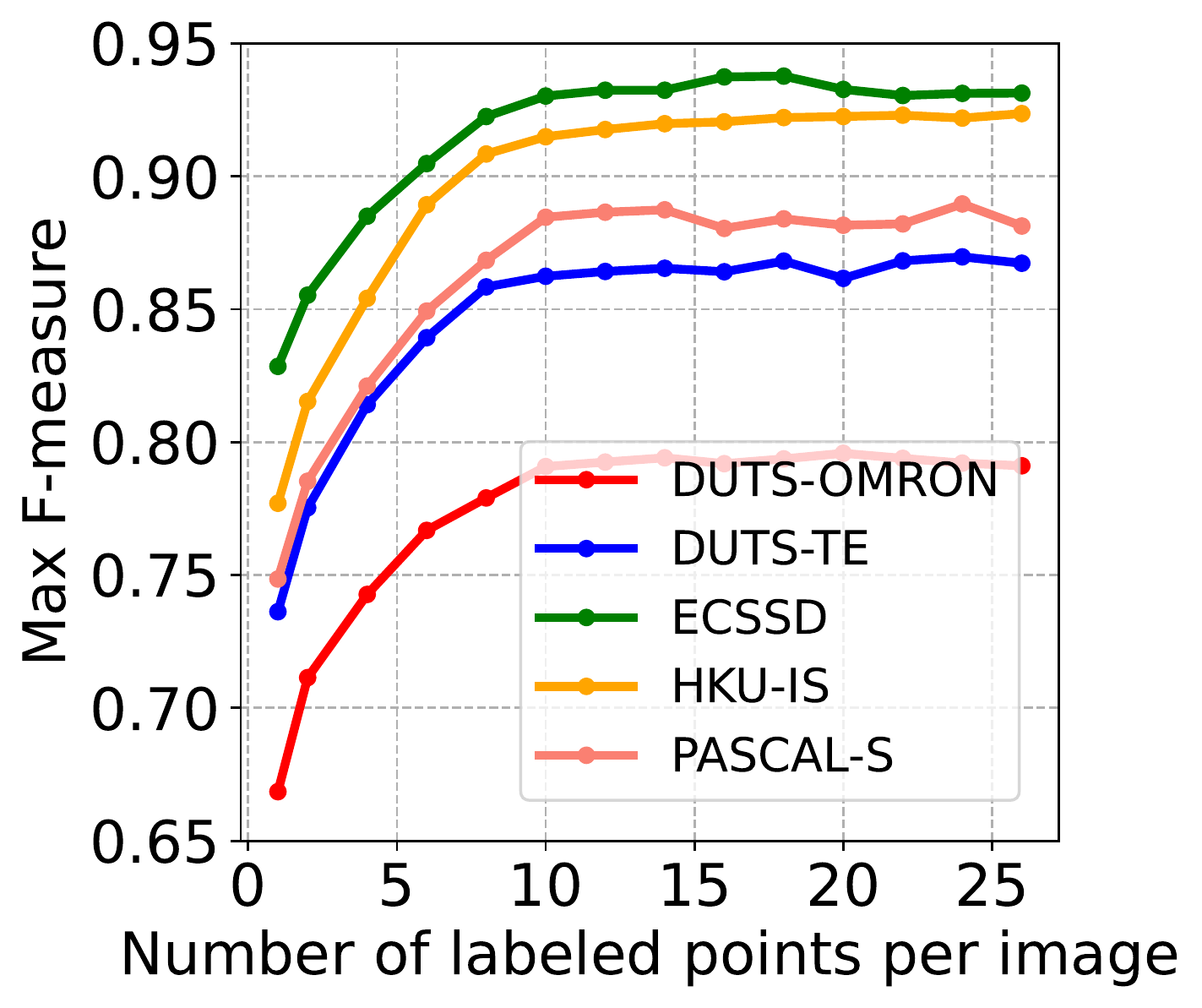}\vspace{-0.5em}
     \caption{}
     \label{fig:pionts_per_image1}
   \end{subfigure}
   \begin{subfigure}{.19\textwidth}
     \centering
     \includegraphics[width=\textwidth]{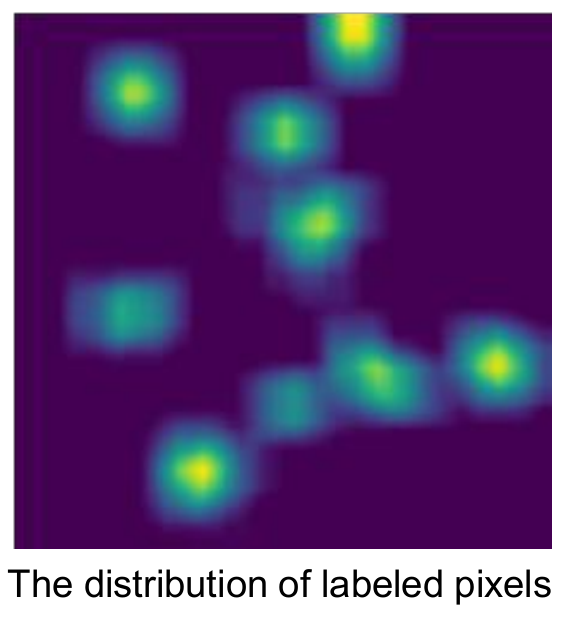}\vspace{-0.8em}
     \caption{}
   \end{subfigure}
     }
   \vspace{-1em}
   \caption{(a) Ablation study on the number of annotated points per image; (b) Our RDS can conquer the oversampling issue, obtaining a set of diverse labeled points.}
   \label{fig:pionts_per_image}
\end{figure}

\noindent{\textbf{Effectiveness of Superpixel (SP).}} Finally, we evaluate the effect using SP selection, forming our complete ATAL algorithm (model \textcircled{5}). Compared to model \textcircled{4}, superpixel selection achieves better performance than pixel selection. When using superpixel as a basic labeled unit, the total labeled pixels account for $2.2\%$ of the DUTS-TR.

\section{Conclusions}

In this paper, we surprisingly find that there is a sparse point labeled dataset where saliency models trained on it can achieve equivalent performance when trained on the densely annotated dataset. As far as we know, this is the first work that gives empirical evidence to the existence of such a sparse point labeled dataset. Our results suggest that sparse point supervision has promising to replace the pixel-wise data in the feature. Besides, we also present an effective and efficient adversarial trajectory-ensemble active learning to identify such a sparse point labeled dataset.
We hope our work encourages further research into the promising use of sparse point-level annotation for image understanding.

\section{Acknowledgments}
This research is supported in part by the National Natural Science Foundation of China (No. 62172437 and 62172246), the Open Project
Program of State Key Laboratory of Virtual Reality Technology and Systems (VRLAB2021A05), the Youth Innovation and Technology
Support Plan of Colleges and Universities in Shandong Province (2021KJ062), and the Science and Technology Innovation Committee of
Shenzhen Municipality (No. JCYJ20210324131800002 and RCBS20210609103820029).

\bibliography{aaai22.bib}


\end{document}